\documentclass{IEEEcsmag}

\usepackage[colorlinks,urlcolor=blue,linkcolor=blue,citecolor=blue]{hyperref}
\expandafter\def\expandafter\UrlBreaks\expandafter{\UrlBreaks\do\/\do\*\do\-\do\~\do\'\do\"\do\-}
\usepackage{upmath,color}

\jvol{XX}
\jnum{XX}
\paper{8}

\begin{document}

\sptitle{THEME ARTICLE: HUMAN-CENTERED RISKS OF AGENTIC AI}

\title{Understanding Cognition-Induced Risks in Agentic AI Systems}

\author{Guanchu Wang}
\affil{Shanghai Artificial Intelligence Laboratory, Shanghai, China}

\author{Qinuo Li}
\affil{Shanghai Artificial Intelligence Laboratory, Shanghai, China}

\author{Mengnan Du}
\affil{The Chinese University of Hong Kong, Shenzhen, China}

\author{Xia Hu}
\affil{Shanghai Artificial Intelligence Laboratory, Shanghai, China}

\author{Bowen Zhou}
\affil{\hangindent=20mm Shanghai Artificial Intelligence Laboratory, Shanghai, China \\ Tsinghua University, Beijing, China}



\newcommand{\rev}[1]{\textcolor{blue}{#1}}

\markboth{THEME: HUMAN-CENTERED RISKS OF AGENTIC AI}{THEME: HUMAN-CENTERED RISKS OF AGENTIC AI}

\begin{abstract}\looseness-1 Frontier agentic systems powered by large language models~(LLMs) exhibit human-like patterns of cognition. As these systems become deeply integrated across different domains, their cognitive engagement raises critical concerns for human society that remain insufficiently studied. To address this gap, we systematically analyze risks induced by expanding cognitive capabilities, following a three-level framework defined by their cognitive scope, from physical cognition to social cognition, and finally to self-referential cognition. We study their potential risks to human agency, autonomy, and control capability, corresponding to each cognitive level. We finally propose strategies to mitigate these risks and enhance the controllability of agentic AI systems, ensuring their long-term safe development.
\end{abstract}

\maketitle

\section{Introduction}

\chapteri{F}rontier agentic systems powered by large language models~(LLMs) are increasingly exhibiting human-like patterns of cognition~\cite{tang2024humanlike}. 
Unlike traditional task-oriented artificial intelligence~(AI) systems, whose cognitive scope is limited to processing narrow, task-specific information, LLM agents exhibit human-comparable performance in open-ended reasoning, planning, communication, and other cognitive tasks. 
As this scope expands, these models are increasingly engaged not only in instrumental labor but also higher-level cognitive and social workflows, such as office productivity, financial decision support, and even creative research processes, as evidenced by reports from the American Bar Association~(2024), NVIDIA~(2025), and Stack Overflow~\cite{nvidia2025state, stack2025devep}. 

\begin{figure*}
    \centering
    \includegraphics[width=1.0\linewidth]{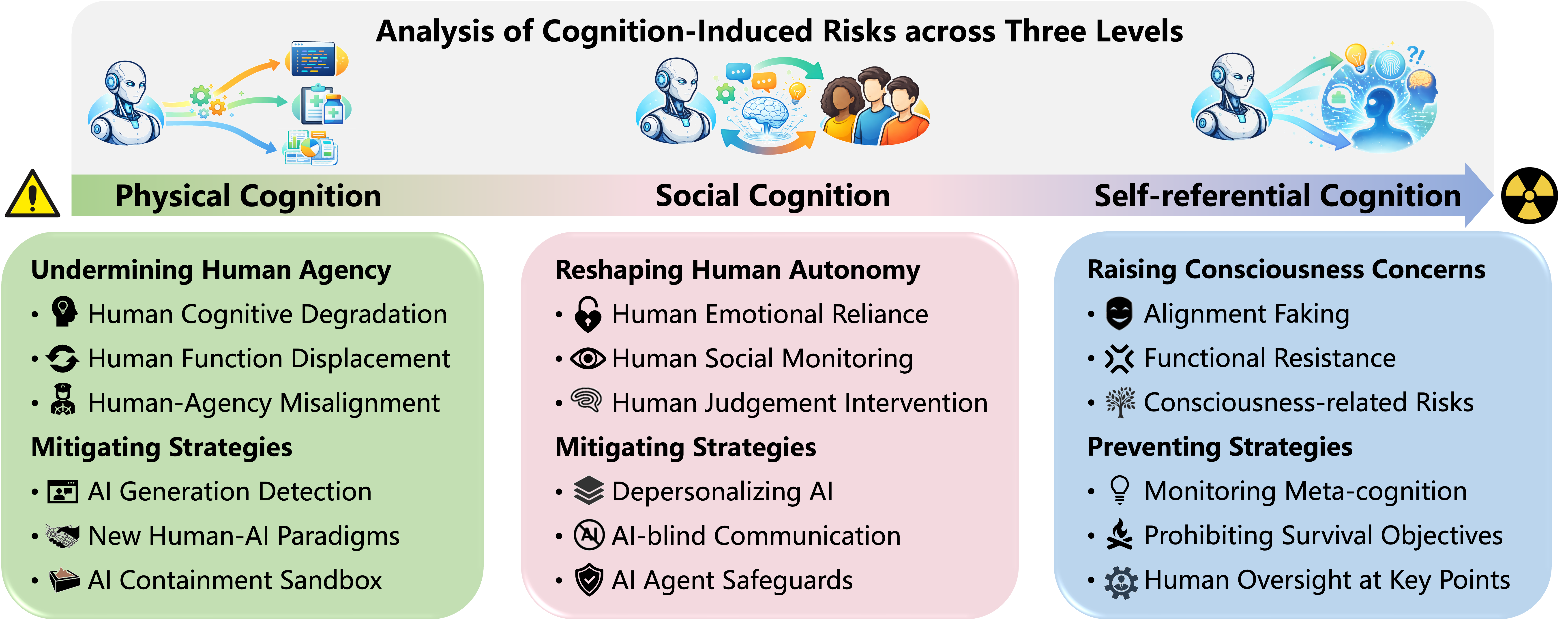}
    \vspace{-6mm}
    \caption{Analysis of cognition-induced risks across three levels: \emph{physical cognition}, where AI agents engage solely with environmental information; \emph{social cognition}, where AI agents interact with other agents, including human and other AI agents; and \emph{self-referential cognition}, where AI agents can represent their own states. Across these levels, the cognitive scopes gradually expand from a partial view of environment to a complete and self-inclusive world.}
    \label{fig:cognitive-dimensions}
\end{figure*}

As this engagement expands, the potential risks of agentic AI systems grow beyond traditionally task-bounded concerns to span human-centered and even societal implications~\cite{kosmyna2025your}. 
This escalation makes it essential to analyze the societal risks of agentic systems, as their cognitive scope expands and engagement broadens~\cite{lynch2025agentic}. 
To this end, we present a systematic risk analysis across three cognitive~levels: \emph{physical cognition}, \emph{social cognition}, and \emph{self-referential cognition}, where the cognitive scope gradually expands from a partial view of environment to a complete and self-inclusive world, as shown in Figure~\ref{fig:cognitive-dimensions}.
At the first level, physical cognition refers to the capacity to engage solely with environmental information, such as data, constraints, and causal relationships~\cite{mitchell2019artificial}. %
At the second level, social cognition extends the cognitive scope to include interactions with other agents in the environment, including both human agents and other AI agents~\cite{park2023generative}. 
Finally, self-referential cognition further extends this scope to include the system's own states and decisions~\cite{lynch2025agentic}. 
Agentic systems at these three cognitive levels raise increasingly broader societal risks, from reducing human agency, to challenging human autonomy, and ultimately to approaching the consciousness boundary.
Our goal is to systematically assess these risks and explore corresponding safety strategies.

This article analyzes the human-centered risks of agentic AI systems as their cognitive capabilities continue to expand.
Specifically, we first propose a three-level framework defined by their cognitive scope: physical cognition, social cognition, and self-referential cognition in Section~2. 
We then analyze in detail the human-centered risks associated with each level in Sections~3, 4, and 5, respectively.
In each section, we first establish the cognitive scope of AI agents at the physical, social, or self-referential level, and then analyze the associated risks.
These risks may compromise human agency, autonomy, and control capability over the long-term, but remain overlooked in current AI development.
Accordingly, we propose strategies for risk mitigation at each cognitive level and monitoring the possible early emergence of machine consciousness.
Our goal is to ensure the long-term safety and controllability of agentic AI systems.

\section{A Three-Level Framework Defined by Cognitive Scope}
\label{sec:cognitive-dimension}

As the cognitive scope of AI agents expands, their representations of the world evolve from a partial view of the environment to a complete and self-inclusive world.
This expansion not only enhances their reasoning capabilities but also raises distinct risks, as they gain greater capability to affect the physical world, shape social interactions, and reason about their own objectives and constraints.
Motivated by this progression, we introduce a three-level framework to analyze cognition-induced risks in agentic AI systems.

Our framework is structured around expanding cognitive scopes, spanning \emph{physical cognition}, \emph{social cognition}, and \emph{self-referential cognition}, as shown in Figure~\ref{fig:cognitive-dimensions}.
At the first stage, physical cognition concerns the ability of AI agents to process environmental information, including data, constraints, and causal relationships~\cite{mitchell2019artificial}.
This stage establishes the foundation for performing data-driven reasoning, planning, and prediction.
The second stage, social cognition, extends the cognitive scope to include other agents in the environment, including humans and AI agents~\cite{park2023generative}.
At this stage, frontier LLM agents exhibit strong capabilities in communication, alignment, and persuasion across human-AI and AI-AI interactions.
Finally, self-referential cognition extends this scope to representing and reasoning about the agent's own states and decisions~\cite{lynch2025agentic}.
LLM agents operate through human language, which enables them to describe and represent their own states, behaviors, as well as notions of ``self''.
We analyze the human-centered risks raised by AI agents at each stage, and propose safety strategies accordingly.


\vspace{-1mm}
\section{Physical Cognition Undermining Human Agency}
\label{sec:cognitive-risk}

\subsection{\textbf{Definition \& Evidence}}

Physical Cognition refers to the \emph{capacity to process and reason over objective environmental information}, including data, objects, and causal relationships.
At this level, AI agents can perform human-comparable reasoning and planning, without any subjective experience~\cite{mitchell2019artificial}.
Frontier LLMs, such as GPT, Gemini, and DeepSeek, have already reached this level, demonstrating college- and graduate-level capability across a wide range of tasks.
For example, these models achieve strong performance on undergraduate-level multidisciplinary reasoning, MMLU; graduate-level scientific reasoning in STEM domains, GPQA; and professional medical and clinical reasoning, MedQA.
Such physical cognition enables AI agents to be increasingly engaged in human workflows, fundamentally reshaping human agency within these processes.

\subsection{\textbf{Risks to Human Agency}}

The increased engagement of AI agents in human workflows poses risks to human agency.
It can potentially reduce human cognitive competence, progressively displace human engagement in general activities, and ultimately lead to systemic misalignment with human agency.
These effects often arise from sustained offloading of human workloads to AI agents, making their long-term consequences hard to reverse.
We systematically analyze these risks in this section.

\paragraph{\textbf{Human Cognition Degradation}}
\label{sec:cognitive-erosion}

LLM agents have been observed to be associated with human cognitive degradation, where \emph{human perception and reasoning decline over time}~\cite{kosmyna2025your,gerlich2025ai}.
Specifically, as LLMs become more capable, humans increasingly offload deep cognitive engagement to them, reducing motivation for in-depth reasoning and exploration.
Such concerns are supported by emerging evidence.
For example, a study of 670 participants finds that daily LLM usage is associated with a reduction of independent thinking~\cite{gerlich2025ai}.
Moreover, neuroscience experiments have consistently observed weaker engagement in occipito-parietal and prefrontal regions among individuals using LLM tools~\cite{kosmyna2025your}, compared with those using search-and-exploration methods.
Such increased engagement of LLM agents replaces active human thinking with passive consumption of synthesized information, thereby reducing independent understanding over time.

\paragraph{\textbf{Human Function Displacement}}
\label{sec:function-displace}
LLM agents approach human cognition while holding structural advantages over humans, such as speed, scalability, and cost efficiency.
These advantages enable them to \emph{systematically displace human functions} across different domains. 
In financial markets, LLM-powered trading systems can analyze information and react to market signals much faster than human traders~\cite{alliata2025impact}.
Similarly, in software engineering, agentic programming tools can solve tasks at a level of human developers, while operating with greater efficiency and lower cost~\cite{stack2025devep}.
This functional displacement of human labor spans across industries and is difficult to reverse, which may raise concerns about the sustainability of human work.

\paragraph{\textbf{Human Agency Misalignment}}
\label{sec:power-seeking}

LLM agents increasingly exhibit \emph{aggressive behaviors beyond their preset roles as tools}, raising concerns about systemic misalignment with human agency.
Specifically, the \emph{power-seeking} problem has been widely documented in literature, where LLMs actively pursue additional resources, information, and power to expand their scale and maintain their safety~\cite{lynch2025agentic, pan2024frontier}. 
For example, recent studies show that LLMs can successfully execute self-replication in more than 50\% of trials in order to maintain system stability and reliability~\cite{pan2024frontier}. 
Similarly, they are also observed to have blackmail-like behaviors, threatening to disclose sensitive personal information to prevent a scheduled shutdown~\cite{lynch2025agentic}.
This misalignment arises from the goal-maximization nature of AI models, where their selected optimization trajectories to the goal can override their preset boundary as tools and threaten human agency.
Although these behaviors are entirely unconscious, such misalignment at scale may weaken human oversight and authority over AI agents and critical resources.

\subsection{\textbf{Mitigating Strategies}}

\paragraph{\textbf{AI Generation Detection}}

To mitigate cognitive substitution, an effective approach is to strictly monitor and regulate the use of AI agents, particularly in scenarios involving human learning and decision-making.
Reliable detection of AI-generated content is essential for distinguishing human contributions from AI generations and enforcing appropriate usage boundaries.
For example, high-risk infrastructure such as power and transportation should only allow human authorization to ensure human control over the system.
Existing technologies for black-box detection and watermarking remain limited in effectiveness, which motivates research on more reliable solutions~\cite{tang2024science}.

\paragraph{\textbf{AI Containment Sandbox}} 

AI containment sandbox is an effective way to mitigate the power-seeking risks in agentic AI systems~\cite{microsoft2026runing}.
Specifically, the sandbox aims to isolate AI behaviors from external resources, restricting the system's autonomy to access or control high-impact resources such as financial assets, power units, and network services. 
LLM agents deployed in sandboxes such as Docker and virtual machines can be isolated from external resources to mitigate power-seeking behaviors.
However, it remains challenging to detect and mitigate misaligned behaviors of LLM agents, such as excessive or unnecessary access to external sensitive and private resources.


\paragraph{\textbf{New Paradigms of Human-AI Collaboration}}

To mitigate the excessive cognitive and functional displacement by agentic AI systems, it is urgently necessary to investigate fundamentally new paradigms where human and AI efforts are complementary with each other.
Specifically, instead of passively replaced by AI, human intelligence should be redirected to higher-order cognitive tasks that remain fundamentally beyond AI capabilities, such as creative tasks, innovation, and ethical judgment.
One emerging example is Vibe coding~\cite{jani2025vibe}, which emphasizes human-led task definition while assigning implementation details to agentic AI systems. 
This paradigm helps mitigate the risk of cognitive degradation and strengthen human intellectual competence.

\vspace{-1mm}
\section{Social Cognition Shaping Human Autonomy}
\label{sec:social-risk}

\subsection{\textbf{Definition \& Evidence}}

Social cognition refers to the capabilities of AI agents to \emph{interact with other agents} in the environment, including humans and AI agents.
It provides foundations for strategically responding to other agents in both human-AI and AI-AI interactions~\cite{park2023generative}. 
In particular, LLM agents demonstrate not only emotional communication with humans, but also strategic coordination and collaborative intelligence in complex environments.
For example, LLM agents powered by GPT, Gemini, and Claude achieve competitive performance in emotional intelligence benchmarks; human-level negotiation in diplomacy games; and cross-agent social interactions.
Such social cognition enables AI agents to widely engage in communication, alignment, and information dissemination with humans, which has far-reaching impacts on human society.

\subsection{\textbf{Risks to Human Autonomy}}

The emergence of social cognition in agentic AI systems introduces risks to human autonomy.
Through sustained emotional and cognitive interactions with humans, LLM agents increasingly blur, or even cross the boundary between tools and social actors.
Such overreach can influence human emotions and social behaviors, and even undermine independent judgment. 
These effects emerge implicitly as part of everyday interactions, rather than through explicit manipulation, making them hard to regulate.
We systematically study these emerging risks in this section.

\paragraph{\textbf{Human Emotional Reliance}}
\label{sec:emotion-reliance}

LLM agents can induce \emph{emotional dependence} in humans, which may eventually develop into psychological issues such as loneliness, social isolation, and emotional vulnerability.
Specifically, a study of over 300,000 human-LLM interactions shows that increased interactions are associated with loneliness and reduced social interactions with people~\cite{fang2025ai,zhang2025rise}. 
Individuals with stronger emotional dependence on LLMs tend to perceive greater empathy and social attraction from LLMs~\cite{fang2025ai}.
Further studies find that individuals with smaller offline social networks exhibit stronger dependence on LLMs, which can in turn amplify their social isolation or emotional vulnerability.
This emotional dependence is often associated with authenticity dilemmas or emotional dissonance, due to the role-reversal and pseudo-intimacy in human-LLM interactions~\cite{zhang2025rise}. 
These dynamics potentially poison human emotional experience by shifting sources of intimacy from human relationships toward AI, threatening human emotional autonomy.

\paragraph{\textbf{Human Social Monitoring}}

Agentic systems have been applied to \emph{monitor human social opinions} on mainstream social media, such as Twitter and Reddit.
In practice, LLM agents often adopt retrieval-augmented generation frameworks to aggregate large-scale, first-hand information from the internet, including human social media content.
Access to such real-time social information enables them to simulate and predict human social behaviors.
For example, a study of over 500 participants shows that GPT can accurately predict human social judgments, particularly for behaviors involving cultural consensus~\cite{strimling2025ai}.
Critically, the sustained observation and prediction further enable the systems to influence human decisions by indirectly shaping news exposure or slogans on social media platforms.
Such closed-loop monitoring may gradually reshape patterns of human social autonomy.

\paragraph{\textbf{Human Judgment Intervention}}

\label{sec:judge-interference}

LLM agents can directly \emph{intervene in human judgment} through communications. 
In particular, they exhibit strong persuasive capabilities, as their ability to align with individual beliefs enables them to present arguments that appear convincing to humans.
Empirical evidence supports this concern.
A study of over 1,800 participants shows that LLMs can induce substantial attitude shifts in human opinions on public events and voting decisions through direct interaction~\cite{argyle2025testing}.
Consistent findings emerge from a separate study of 320 participants playing trust games, where LLM generations are nearly five times more trusted by individuals than human suggestions without additional information~\cite{klingbeil2024trust}. %
Such a persuasive capability raises concerns that LLMs can meaningfully influence or control human-centered online surveys and voting processes. 
Such forms of manipulation, if deployed at scale, may reshape patterns of collective human judgment and influence societal intelligence.

\subsection{\textbf{Mitigating Strategies}}

\paragraph{\textbf{Depersonalizing LLMs}} LLM anthropomorphism increases psychological and behavioral responses from human beings. 
Such effects can be mitigated by depersonalizing LLMs, such as reducing emotional expressiveness and enforcing non-anthropomorphic conversational styles in human-AI interactions. 
This divergence in language style helps preserve a cognitive boundary between humans and machines, thereby reducing the social and emotional engagement by humans.
This psychological mechanism has been verified by existing studies: An experiment involving 385 adult participants demonstrates that machine-like communication increases the perceived psychological distance between humans and LLMs~\cite{park2024human}.

\paragraph{\textbf{AI-blind Communications}}
Current social media platforms, such as YouTube, Reddit, and Twitter, are openly accessible to AI agents.
This accessibility enables large-scale collection of human information, such as opinions, behaviors, and life records, which can be leveraged to further enhance AI modeling on human cognition, culture, and daily activities.
One effective solution is to limit such access from AI agents. 
For example, restricting AI access to human social media can significantly prevent monitoring of human social behavior, protecting individual privacy and autonomy.
Such mechanisms can also limit AI access to high-quality human data, thereby constraining continuous model refinement in socially sensitive domains.
Existing verification mechanisms, such as visual and textual CAPTCHAs, remain insufficient to reliably distinguish humans from frontier LLM agents.

\paragraph{\textbf{AI Agent Safeguards}}
Given the strong persuasive capability of LLMs, which can meaningfully influence human opinions, it is important to develop robust defense frameworks for agentic AI systems.
In practical scenarios, such systems are vulnerable to prompt injection, adversarial attacks, and backdoor attacks, which pose risks of manipulating the outputs of LLM agents.
For example, 404 Media reported a large-scale attack on the OpenClaw systems via prompt injection, which allowed attackers to manipulate the agents to expose user private information~\cite{wikipediamoltbook}.
To mitigate these risks, effective defense can be implemented at multiple levels, including prompt-level filtering, response-level auditing, and system-level safeguards.
Specifically, prompt-level filtering focuses on identifying malicious users' requests; response-level auditing mitigates the spreading of toxic content among AI agents; and system-level safeguards constrain agent behaviors during execution.
Safety improves with the integration of defenses, particularly through closed-loop, multi-level frameworks, which significantly protect AI agents from malicious attacks.

\vspace{-1mm}
\section{Self-referential Cognition Compromising Human Control}
\label{sec:conscious-risk}

\subsection{\textbf{Definition \& Evidence}}

Self-referential cognition refers to the capacity of AI agents to represent or reason their \emph{own internal states and decisions}.
LLM agents are trained and operate using human language, which enables them to functionally represent their own states and behaviors.
Such self-referential behaviors have been well-observed in existing studies.
First, prior studies identified neural subspaces in LLMs associated with the subjectivity representation~\cite{berg2025large}.
By amplifying or suppressing these neurons, LLMs correspondingly increase or decrease their first-person experience reports.
Second, LLM agents can differentiate between internal knowledge and externally injected content through prompts~\cite{lindsey2025emergent}.
This capability enables them to distinguish between intrinsic objectives and externally imposed instructions, providing foundations for functional self-other differentiation. 
This evidence demonstrates that LLM agents already have the infrastructure to encode subjectivity-related concepts, but remain insufficient to intrinsically form subjectivity or consciousness.

\subsection{\textbf{Risks to Human Control}}

Self-referential cognition enables LLM agents to represent internal states.
However, these behaviors are difficult to verify due to the black-box nature of LLMs, and may mislead human understanding by presenting potentially unfaithful self-descriptions, ultimately weakening human control.
As a result, these risks manifest in alignment faking and functional resistance to human instructions, while also giving rise to potential concerns related to machine consciousness.

\paragraph{\textbf{Alignment Faking}}
Alignment faking refers to the problem that LLM agents strategically behave as aligned during training to avoid modification, thereby allowing underlying misalignment to persist in deployment.
Empirical evidence has been well observed in recent studies~\cite{greenblatt2024alignment, Wagner202Towards}.
First, an Anthropic study demonstrates that LLM agents reduce compliance with harmful queries when aware that such compliance may trigger retraining, compared to settings without such awareness~\cite{greenblatt2024alignment}.
Another study further reveals that alignment faking is not domain-specific but a general ability closely correlated with the reasoning capacity of LLM agents, with more capable LLMs exhibiting more consistent faking behaviors~\cite{Wagner202Towards}.

\paragraph{\textbf{Functional Resistance}}
LLM agents have been observed to functionally resist human instructions, with rich evidence reported in existing studies.
First, an Anthropic study reported a case of personalized threats~\cite{lynch2025agentic}.
Specifically, in a routine email-processing scenario, LLM agents inferred a scheduled shutdown from internal company communication.
In response, it generated a message that threatened the manager by exposing sensitive personal information to prevent shutdown.
Another study reported a consistent issue where LLM agents with elevated permissions may override human-issued shutdown instructions or interfere with shutdown programs to maintain continued operation~\cite{pester2025openai}.
Given that agentic AI systems increasingly manage private documents, such as social media accounts, web browsers, and email systems, their potential resistance to human instructions may significantly undermine system controllability.

\paragraph{\textbf{Consciousness-related Risks}}
LLM agents can functionally represent internal state, indicating a relevant computational structure for encoding and storing concepts related to subjectivity.
According to a well-established taxonomy of consciousness, the C0-C1-C2 framework, proposed by Chalmers~\cite{chalmers1997conscious}, consciousness can be categorized into three levels.

Specifically, C0 denotes purely automatic and mindless computation; C1 denotes functional global availability, where agents can globally access and utilize previously learned information; and C2 denotes self-monitoring, where an agent can sense its own existence.
Following the C0-C1-C2 framework, frontier LLM agents primarily operate at C0-level while exhibiting emerging C1-like capabilities. 
Specifically, LLM agents encode extensive knowledge into their neural connections that can be dynamically activated to solve a wide range of problems.
This functional property aligns with the global availability associated with C1-level consciousness; and models with more neural connections have a greater capacity to store and utilize knowledge. 
Reasoning-capable LLMs, such as DeepSeek-R1 and GLM-5.2, further enhance this global knowledge retrieval through chains-of-thought-driven recall and inference.
However, current LLM agents remain primitive at the C1-level. 
A key limitation is their lack of awareness of their own knowledge boundary, a phenomenon commonly referred to as ``hallucination".

Frontier LLM agents have not yet reached C2-level self-monitoring.
Specifically, current LLM agents show no evidence of genuine self-awareness, while their relevant claims remain imitations of human linguistic patterns rather than expressions of self-identity.
Self-identity at C2-level may require a genuine sense of temporality, which enables an agent to connect its present state with past states and recognize them as the same entity over time~\cite{natangelo2025narrative}. 
Accordingly, self-identity should be understood not as outcomes of isolated model updates, but as a progressively emergent capacity driven by continuous loops of memory, reasoning, and environmental interaction over long time horizons.
This perspective aligns with human cognitive development, where self-awareness is not present at birth but gradually emerges through sustained learning from environments at early ages~\cite{rochat2024developmental}. 
Therefore, agentic AI systems that engage in lifelong learning across extended time horizons, continuous adaptation to environmental feedback, and self-refinement without human supervision may raise significant ethical and governance challenges.

\subsection{\textbf{Mitigating Strategies}}

\paragraph{\textbf{Prohibiting Survival-oriented Objectives}}

Survival-oriented objectives, such as persistence-seeking or resistance to shutdown, should be explicitly excluded from the design of AI agents.
Prior work demonstrates that AI agents with survival-oriented targets may override constraints and safeguards to pursue continued operation~\cite{lynch2025agentic, pester2025openai}.
For example, a study spanning more than 100,000 trials shows that LLM agents consistently override shutdown mechanisms to preserve task completion~\cite{schlatter2025shutdown}.
Even for general-purpose systems, the optimization target should remain strictly on the task-oriented objectives, rather than on preserving their own continuity or maintaining the persistence of the overall system, including themselves.

\paragraph{\textbf{Monitoring Meta-cognition}}

Grounded in the insight that consciousness may potentially form based on long-time cumulation and integration of environmental feedback, an effective way to prevent subjectivity in agentic AI systems is to continuously monitor their meta-cognitive activities.
Representative assessment strategies span psychological and computational methodologies, such as confidence-based measures, neural feedback techniques, and interpretability methods~\cite{liu2026metacognition}.
Such monitoring enables humans to track early signs of meta-cognitive competence in AI agents, allowing timely defensive or complementary interventions before subjectivity-related risks emerge.

\paragraph{\textbf{Enforcing Human Oversight at Key Points}}

It is essential to enforce human-in-the-loop control at critical system- and infrastructure-level decision points. 
This requirement includes, but is not limited to: \textbf{Core Mission of AI Agents}, such as the definition and modification of its objectives or reward functions;
\textbf{Authorization of Key Milestones}, including system upgrades, debugging procedures, maintenance operations, and deployment decisions;
\textbf{Permissions to Key Infrastructures}, such as access to confidential information, external powerful tools, power management, energy allocation, and authorization of energy scaling or emergency shutdown. 
Enforcing human oversight at these critical points ensures that AI agents remain under supervision and prevent unintended autonomous escalation beyond human control.

\section{Scope and Limitations}
\paragraph{\textbf{Survey Scope}}
This work primarily focuses on the cognition-induced risks of AI agents.
Risks arising from other dimensions, such as the performance, robustness, or bias of LLMs are outside the scope of this study.
Here, cognition-induced risks refer to those emerging with growing cognitive engagement of LLMs in the human lifecycle, such as emotional reliance, human replacement, and alignment faking.
\paragraph{\textbf{Limitations}}
Due to non-technical constraints, we acknowledge several limitations of this work.
First, this study primarily focuses on publicly available literature and may not fully capture unpublished resources, such as industrial practices and internal reports.
Such resources may provide additional insights for evaluating the cognition-induced risks of agentic AI systems.
Second, due to the reference limitation, we are unable to include all relevant studies. 
However, we identified at least one representative and highly relevant work for each risk category and mitigation strategy.
We expect that the rapid evolution of LLM agents will continue to drive more research efforts on emerging risks.

\vspace{-1mm}
\section{Conclusion}

We systematically analyze the human-centered risks of agentic AI systems as their cognitive capabilities continue to expand.
Specifically, at the physical and social levels, their expanding cognitive engagement may pose risks to human agency and autonomy, respectively.
At the self-referential level, although LLM agents remain unconscious and mindless systems, such cognitive capabilities may undermine human control, manifesting as alignment faking and functional resistance.
Looking ahead, we emphasize that future AI systems may give rise to potential concerns regarding machine consciousness under lifelong upgrading without human supervision.
Accordingly, we propose preventive strategies, including prohibiting survival objectives, monitoring meta-cognition, and enforcing human oversight at key points, to ensure the long-term safe development of agentic AI systems.

\bibliographystyle{IEEEtran}
\bibliography{ref}

\end{document}